\documentclass[lettersize,journal]{IEEEtran}
\usepackage{amsmath,amsfonts}
\usepackage{algorithmic}
\usepackage{algorithm}
\usepackage{array}
\usepackage[caption=false,font=normalsize,labelfont=sf,textfont=sf]{subfig}
\usepackage{textcomp}
\usepackage{stfloats}
\usepackage{url}
\usepackage{verbatim}
\usepackage{graphicx}
\usepackage{cite}
\begin{document}

\title{Fine-Grained 3D Facial Reconstruction for Micro-Expressions}

\author{Che Sun, Xinjie Zhang, RuiGao, Xu Chen, Yuwei Wu, Yunde Jia }
      %~\IEEEmembership{Staff,~IEEE,}
        % <-this % stops a space



\maketitle

\begin{abstract}
% Recent advances in 3D facial expression reconstruction have achieved impressive results for macro-expressions, yet the reconstruction of micro-expressions remains unexplored. This novel task is particularly challenging due to the subtle, transient, and low-intensity nature of micro-expressions, which are difficult to capture and model especially under small-scale training data. In this paper, we propose a dynamic-aware progressive reconstruction method for fine-grained 3D facial micro-expression modeling. Our method encodes global facial dynamics, and subsequently applies localized and detail-preserving refinements for 3D micro-expressions, constituting a coarse-to-fine framework for dynamic modeling. Specifically, we devise a plug-and-play dynamic-encoded module to capture micro-expression facial actions, allowing it to leverage prior knowledge from abundant macro-expression data to mitigate the scarcity of micro-expression data. Subsequently, a dynamic-guided mesh deformation module is designed to leverage dense optical flow and sparse landmark cues to adaptively refine fine-grained facial micro-expression without compromising global 3D geometry. Extensive experiments on micro-expression datasets demonstrate that our method consistently outperforms state-of-the-art methods in both geometric accuracy and perceptual detail. 
%To our knowledge, this represents the first dedicated method for high-fidelity 3D reconstruction of facial micro-expressions, offering new possibilities for affective computing and behavior analysis.

Recent advances in 3D facial expression reconstruction have demonstrated remarkable performance in capturing macro-expressions, yet very few studies have focused on reconstructing micro-expressions. This novel task is particularly challenging due to the subtle, transient, and low-intensity nature of micro-expressions, which complicates the extraction of stable and discriminative features essential for accurate reconstruction. In this paper, we propose a fine-grained micro-expression reconstruction method that integrates global dynamic features capturing stable facial motion patterns with locally-enriched features incorporating multiple informative cues from 2D motion, facial priors, and 3D facial geometry. %Our method fuses multi-model cues in a gloal-to-lcoal step to extract  stable dynamic feature for micro-expressions, enabling a guided deformation module to progressively impart detail-preserving refinements to the global geometry
 %Extensive experiments on benchmark micro-expression datasets demonstrate that our method consistently surpasses state-of-the-art approaches in both geometric accuracy and perceptual detail.
Specifically, we devise a dynamic-encoded module to extract micro-expression features  for global  facial motion, allowing it to leverage prior knowledge from abundant macro-expression data to mitigate the scarcity of micro-expression data. Subsequently, a dynamic-guided mesh deformation module is introduced for extracting aggregated local features from dense optical flow, sparse landmarks, and facial geometry, which adaptively refines fine-grained facial micro-expressions without compromising global 3D geometry. Extensive experiments on micro-expression datasets demonstrate the effectiveness of our method.  %that our method outperforms state-of-the-art methods. 
%Extensive experiments on micro-expression datasets demonstrate that our method consistently outperforms state-of-the-art methods in both geometric accuracy and perceptual detail. 
\end{abstract}

% Recent advances in 3D facial expression reconstruction have achieved impressive results for macro-expressions, yet the reconstruction of micro-expressions remains largely unexplored. This task is particularly challenging due to the subtle, transient, and low-intensity nature of micro-expressions, which are difficult to capture and model—especially under limited training data. To address this, we propose a novel dynamic-aware progressive framework for high-fidelity 3D facial micro-expression reconstruction. Our method follows a coarse-to-fine pipeline: it first anchors the global expression in a dynamic-aware embedding space, then enhances it with localized, detail-preserving refinements. 
% Specifically, we introduce a dynamic-aware expression embedding within the FLAME parameter space, enabling plug-and-play integration with pre-trained macro-expression models while enhancing dynamic sensitivity to mitigate data scarcity. Subsequently, a dynamic-guided mesh refinement network leverages dense optical flow and sparse landmark cues to adaptively recover fine-grained facial deformations without compromising global geometry. Extensive experiments on synthetic and real micro-expression datasets demonstrate that our approach consistently outperforms state-of-the-art methods in both geometric accuracy and perceptual detail. To the best of our knowledge, this is the first dedicated framework for high-quality 3D micro-expression reconstruction, offering new potential for affective computing and fine-grained behavior analysis.

\section{Introduction}
\label{sec:intro}
% Realistic 3D facial expression reconstruction empowers emotionally intelligent social robots that engage in companionship and caregiving interactions with humans~\cite{KongXYCKZT25,LiuZLZZL25,HuCLWWML24,SuguitanDHH24}. 
Realistic 3D facial expression reconstruction is critical for advancing artificial intelligence systems that rely on perceptual and emotional understanding~\cite{KongXYCKZT25,LiuZLZZL25}, particularly in social robots designed for companionship and caregiving~\cite{HuCLWWML24,SuguitanDHH24}.
% Realistic 3D facial expression reconstruction is critical for advancing artificial intelligence systems that rely on perceptual and emotional understanding ~\cite{KongXYCKZT25,LiuZLZZL25}, underpinning social robots for companionship and caregiving~\cite{HuCLWWML24,SuguitanDHH24}, as well as virtual reality~\cite{LouWNHMWY20}, and embodied agents~\cite{ZhouLHJ23}.
Significant progress has been achieved in reconstructing macro-expressions~\cite{RetsinasFDARBM24, QianKS0GN24}, i.e., long-duration and easily recognizable expressions that convey clear emotions, while few studies have focused on micro-expression reconstruction. Micro-expressions are involuntary, fleeting, and subtle (typically under 0.5 seconds) facial expressions that reveal hidden or suppressed emotions~\cite{ZhaoLLP23}. Their subtle and rapid dynamics make them particularly informative for understanding undercurrent emotion states, while also posing significant challenges for capturing and reconstructing them. In this work, we focus on faithfully reconstructing the fine-grained dynamics of micro-expressions, enhancing the ability of intelligent robots to interpret and simulate subtle human emotions.

Reconstructing 3D facial micro-expressions is non-trivial, because extracting stable and discriminative features is complicated stemming from the unique characteristics of the extremely subtle and low-intensity facial dynamics. The low-intensity signals could be easily dominated by ubiquitous noise, such as illumination changes, head movements, and sensor artifacts, hindering these delicate signals from being encoded into the feature representations. Besides, different micro-expressions often manifest as minute variations within highly overlapping facial regions due to their locally confined nature. This leads to low separability in the feature space, making it difficult to learn discriminative features that can reliably discriminate between subtle yet semantically different affective states.

% Moreover, 
% typically confined to local regions like the mouth or eyes rather than involving global facial motions. This fine-grained and localized nature demands that are both highly sensitive to minute local changes and robust enough to maintain spatiotemporal consistency across the face.

To address the issue, we propose a fine-grained micro-expression reconstruction method that integrates global dynamic features with locally-enriched features of micro-expressions. Our method encodes global dynamic features to capture holistic facial motion patterns, which provides stable spatial-temporal contextual priors to mitigate the influence of noise and enhance overall temporal consistency. We further aggregate multiple cues, including 2D motion patterns, facial semantic priors, and 3D facial geometry, thus obtaining discriminative local features through cross-modal complementarity.  The integration of globally stabilized dynamics with these locally-enriched features collectively facilitates the precise reconstruction of subtle and localized micro-expressions. Specifically, we devise a plug-and-play dynamic-encoded module to extract global facial features from both RGB and optical-flow sequences, effectively leveraging prior knowledge from abundant macro-expression data to alleviate the data scarcity of micro-expressions. A dynamic-guided mesh deformation module, which fuses locally-enriched features from dense optical flow, sparse landmarks, and 3D geometry, is introduced to adaptively refine local facial details while preserving the global facial structure.

Prior to this work, there were few dedicated benchmarks for 3D micro-expression reconstruction. Therefore, we repurpose three high-frame-rate micro-expression recognition datasets of CASME~\cite{yan2013casme}, CASME II~\cite{YanLWZLCF14}, and SAMM~\cite{DavisonLCTY18} for evaluation, because they have proven ability to capture subtle and rapid facial dynamics. Experimental results on the three datasets show the effectiveness of our method.

The main contributions of this work are summarized as follows:
\begin{itemize}
\item To our best knowledge, we are the first to reconstruct fine-grained 3D facial micro-expressions.  Our method encodes global facial dynamics, and subsequently applies localized and detail-preserving refinements for 3D micro-expressions, constituting a coarse-to-fine framework for effective 3D facial reconstruction.

\item We propose a robust feature extraction strategy that integrates global dynamics with locally-enriched multi-modal cues, effectively suppressing noise and enhancing discriminability for subtle micro-expression dynamics. 

% \item We design and develop a novel and highly expressive human-face robot, namely Pengrui, serving as a dedicated  experimental platform for real-world expression imitation to validate the  practical effectiveness of our method. 
\end{itemize}

\IEEEpubidadjcol

\section{Related Work}
\label{sec:relate}

% \subsection{3D Face Reconstruction}
We review the previous works that are closely related to our method in the field of monocular 3D facial reconstruction and head avatar creation from videos. 

\subsection{Monocular 3D Facial Reconstruction}
Monocular 3D facial reconstruction has seen significant advances in recent years, and existing reconstruction methods are broadly categorized into model-free~\cite{RuanZWWW21,ZengP019,SelaRK17} and model-based paradigms~\cite{DanecekBB22,RetsinasFDARBM24,FengFBB21}. Model-free methods typically regress 3D geometry in the form of 3D meshes~\cite{WuRV23,ZengP019}, voxels~\cite{JacksonBAT17}, or implicit representations such as Signed Distance Functions~\cite{ChatziagapiAMS21,YenamandraTBSEC21}.  While these methods exhibit flexibility, they generally rely on large-scale 3D training data, which is synthetically generated using parametric face models. This dependency, however, could influence their reconstruction fidelity, as the quality and diversity of the training data are bounded by the expressiveness of the underlying generative model~\cite{RetsinasFDARBM24,RuanZWWW21,DouSK17}. Differently,  model-based methods fit the low-dimensional parameters of a statistical face model, such as the Basel Face Model (BFM)~\cite{PaysanKARV09} or FLAME~\cite{RetsinasFDARBM24,LiMVEK23}. To circumvent the scarcity of large-scale 3D  training data, recent model-based methods have increasingly shifted towards self-supervised learning frameworks~\cite{RetsinasFDARBM24,LiMVEK23,DanecekBB22,TewariZK0BPT17}. These methods typically employ supervision from landmark re-projection and photometric consistency losses.  Recent occlusion-inpainting methods like OFER~\cite{SelvarajuABADAZ25} and  EIFER~\cite{Buchner0GD25} focus on recovering local dynamics under partial observations by using FLAME models. 

While existing methods have shown impressive capability in reconstructing macro-expressions, they pay little attention to micro-expressions. Differently, our method focuses on the reconstruction of micro-expressions by integrating global dynamics and locally-enriched features to better capture their fine-grained and transient facial dynamics.

% \subsection{Micro-Expression Generation}
\subsection{Head Avatar Creation}
Different from monocular 3D facial reconstruction, the creation of high-fidelity animatable head avatars rely on multi-view video inputs to achieve higher robustness and visual quality~\cite{BaoLCZWZKHJWYZ22,HuSWNSFSSCL17}. Most existing methods often adopt a single representation, such as meshes~\cite{BharadwajZHBA23,GrassalPLRNT22}, point clouds~\cite{WangKCBSZ23,ZhengABCBH22}, or implicit radiance fields~\cite{ChenOB023,GaoZXHGZ22,Yang05637,ZhaoWSZSL24}, to model the entire head. For example, GaussianAvatars~\cite{QianKS0GN24} uses rigged 3D Gaussians driven by a FLAME mesh, allowing high-frequency hair modeling and real-time rendering. Concurrently, GAF~\cite{Tang0KSN25} explores monocular video input for Gaussian-based avatar reconstruction, yet remains constrained by dynamic motion blur and limited-resolution texture details compared to multi-view approaches. Recent works, such as DELTA~\cite{Feng2309} and  MeGA~\cite{WangKSQWBZ25}, find that 3D Gaussians excel at modeling volumetric structures but struggle with fine-grained facial details, and NeRF or mesh-based methods often fail to unify high-fidelity skin and complex hair. Therefore, they turn to hybrid or component-aware representations to address these limitations. 

Existing head avatar creation methods are generally designed for macro-expressions with large head movements, typically relying on multi-view inputs or substantial pose variations. Differently, our work emphasizes the reconstruction of fine-grained micro-expressions from monocular high-frame-rate videos by integrating multi-source cues.

\section{Preliminaries}
\label{sec:preliminaries}
\subsection{Problem Formulation}
Given a monocular video $V=\{I_t \in \mathbb{R}^{H \times W \times 3}\}_{t=1}^T$ with $T$ RGB frames $I_t$, the goal of facial reconstruction is to estimate a detailed 3D facial geometry that faithfully captures both identity characteristics and expressive deformations. The output comprises a sequence of 3D meshes $\{\mathcal{M}_t = (\mathcal{V}_t, \mathcal{F}_t)\}_{t=1}^T$ with vertices $\mathcal{V}_t \in \mathbb{R}^{n_v \times 3}$ and faces $\mathcal{F}_t$ in the $t$-th frame. %, along with associated expression parameters that enable animation and manipulation.

\subsection{Analysis-by-Synthesis Paradigm}
Modern learning-based approaches \cite{RetsinasFDARBM24,SelvarajuABADAZ25} employ an analysis-by-synthesis paradigm, which consists of a face parameter encoder, a 3D face model, and a neural render for reconstructing expressive 3D faces in a self-supervised training manner.  

\noindent\textbf{Encoder:} A deep network $E^{\text{reg}}$ regresses the parameters of a 3D Morphable Model (3DMM) FLAME \cite{LiBBL017}, given by
\begin{equation}
    \beta_t, \psi_t, \theta_t = E^{\text{reg}}(I_t), \label{equ:flame1}
\end{equation}
where $\beta_t$ denotes the shape parameters, $\psi_t$ indicates the expression parameters, and $\theta_t$ indicates the pose parameters. %Due to the focus on improving facial expression reconstruction, it is assumed during training that $\beta_t$ and $\psi_t$ have been pre-trained and remain frozen.

\begin{figure}[t]
    \centering
    \includegraphics[width=1.0\linewidth]{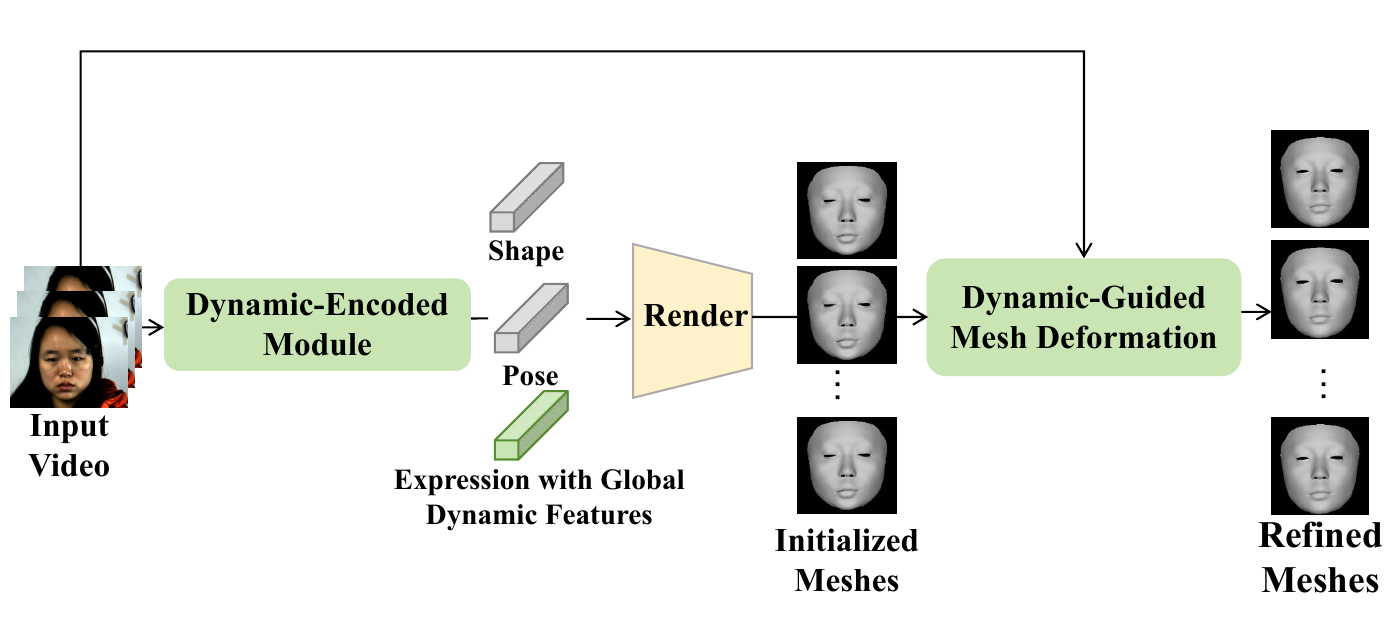}
    \caption{The overall framework of our method.}
    \label{fig:overall}
\end{figure}

\noindent\textbf{3D Face Model:} The FLAME model generates the 3D meshes through
\begin{equation}
    \mathcal{V}_t, \mathcal{M}_t =FLAME(\beta_t, \psi_t, \theta_t), \label{equ:flame2}
\end{equation}
which captures facial shape variations across shapes and expressions through linear blendshapes.

\noindent\textbf{Neural Render:} A neural image-to-image translator $\mathcal{T}$ is introduced to achieve effective analysis-by-neural-synthesis supervision. Given a monochrome rendering of the reconstructed face mesh $S_t = R(\beta_t, \psi_t, \theta_t)$ and a sparsely input image $M(I_t)$ where only $1\%$ of facial pixels are retained, the neural render reconstructs the original image as 
\begin{equation}
    \hat{I}_t = \mathcal{T}(S_t \oplus M(I_t)),\label{equ:render}
\end{equation}
where $\oplus$ denotes concatenation. This formulation forces $\mathcal{T}$ to rely on geometric cues from $S_t$.  %A multi-objective optimization from the reconstruction is used, including photometric $\mathcal{L}^{\text {photo}}$, perceptual $\mathcal{L}^{\text {vgg}}$, landmark $\mathcal{L}^{\text {lmk}}$, and emotion $\mathcal{L}^{\text {emo}}$ losses for reconstruction fidelity. 

\noindent\textbf{Augmented Expression Cycle Path:} Starting from predicted parameters $(\beta_t, \psi_t, \theta_t)$,  the expression $\psi_t$ is replaced with an augmented version $\psi_t^{\text {aug}}$, and  the corresponding image is generated via neural rendering $\hat{I}_t^{\text {aug}} = \mathcal{T}(R(\theta_t, \beta_t, \psi_t^{\text {aug}}) \oplus M(I_t))$. The encoder $E^{\text{reg}}$ must recover the modified expression from the generated image, enforced by the cycle consistency between $E^{\text{reg}}(\hat{I}_t^{\text {aug}})$ and $\psi_t^{\text {aug}}$. 

The analysis-by-synthesis paradigm is effective for macro-expression reconstruction, yet struggles to capture the subtle dynamics of micro-expressions. Due to their low-intensity nature, micro-expressions induce minimal dynamics in expression parameters $\psi_t$ and facial mesh geometry $\mathcal{M}_t$, making it challenging to extract stable and discriminative features from such fine-grained facial dynamics.

% This limitation stems from inadequate feature extraction for such fine-grained facial movements and the scarcity of micro-expression training data.

% This approach mitigates expression collapse and prevents over-compensation between encoder and renderer through diverse expression augmentation including permutation, perturbation, and template injection.

% A multi-objective optimization is used, including photometric, perceptual, landmark, and emotion losses for reconstruction fidelity, alongside cycle-consistency losses for expression augmentation and identity preservation. The losses are optimized alternately between reconstruction and augmentation paths to ensure geometric accuracy and expression diversity.

% A renderer $R$ synthesizes a 2D image from the 3D mesh:
% \begin{equation}
%     \hat{I} = R(\mathcal{V}, \bm{\alpha}, \bm{l})
% \end{equation}
% where $\bm{\alpha}$ represents albedo and $\bm{l}$ lighting parameters.

% \noindent\textbf{Optimization Objective:} The framework is trained by minimizing the discrepancy between input and synthesized images:
% \begin{equation}
%     \mathcal{L}_{total} = \lambda_{photo}\mathcal{L}_{photo}(\hat{I}, I) + \lambda_{lmk}\mathcal{L}_{lmk} + \lambda_{perc}\mathcal{L}_{perc} + \lambda_{reg}\mathcal{L}_{reg}
% \end{equation}
% where $\mathcal{L}_{photo}$ measures photometric consistency, $\mathcal{L}_{lmk}$ enforces landmark alignment, $\mathcal{L}_{perc}$ incorporates perceptual similarity, and $\mathcal{L}_{reg}$ imposes statistical priors on the parameters.

\section{Method}
\label{sec:method}
Our method aims to reconstruct fine-grained 3D facial micro-expressions $\{\mathcal{M}_t = (\mathcal{V}_t, \mathcal{F}_t)\}_{t=1}^T$ from a monocular video $V$ through a coarse-to-fine framework. The overall framework is depicted in Figure~\ref{fig:overall}. The \textit{Dynamic-Encoded} module takes the input video and generates initialized 3D meshes by integrating global dynamic features. The \textit{Dynamic-Guided Mesh Deformation}  module refines the initialized meshes using locally-enriched features to capture subtle dynamic details of  micro-expressions.

\begin{figure}[t]
    \centering
    \includegraphics[width=1.0\linewidth]{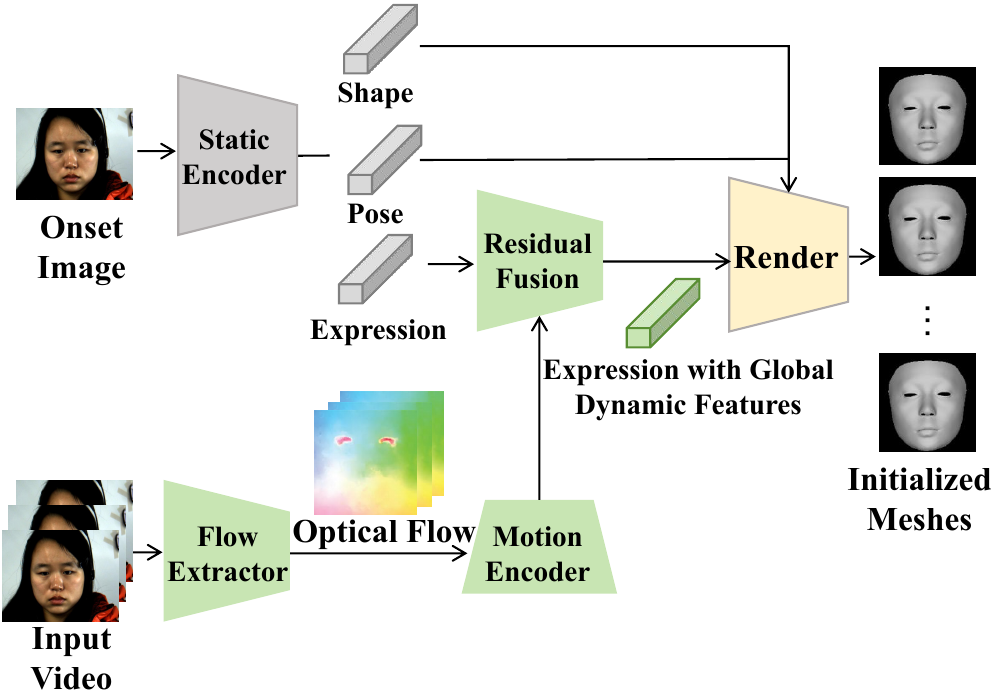}
    \caption{The \textit{dynamic-encoded} module generates initialized 3D facial meshes from an onset image and input video. %The onset image is encoded into shape, pose, and expression parameters. Optical-flow sequences from the input video fuses with expression parameters via the residual fusion to form the expression parameters with global dynamic. Finally, all parameters are rendered into the initialized meshes.
}
    \label{fig:dynamic_encoded}
\end{figure}

\subsection{Dynamic-Encoded Module}
\label{subsec:dynamic_encoded}
The dynamic-encoded module, illustrated in Figure~\ref{fig:dynamic_encoded}, is designed as a plug-and-play component that generates initialized 3D facial meshes $\{\mathcal{M}_t^{\text {init}} = (\mathcal{V}_t^{\text {init}}, \mathcal{F}_t^{\text {init}})\}_{t=1}^T$ by leveraging both static shape  and expression motion to extract global dynamic features. 
%The plug-and-play nature enables seamless integration with existing macro-expression reconstruction methods while specifically addressing the challenges of reconstructing micro-expression dynamics. %This design allows us to transfer prior knowledge from abundant macro-expression data to mitigate the scarcity of micro-expression training samples. %可以作为贡献列举（check一下写法）

The module operates through two parallel streams. (1)  The static encoder extracts static shape $\beta_1$, expression $\psi_1$ and pose $\theta_1$ parameters from the onset image $I^{\text {onset}}$ (i.e., $I_1$) serving as the anchor frame for expression dynamics. We follow the previous reconstruction work of Retsinas et al.~\cite{RetsinasFDARBM24} to construct the encoder by using Eq. (\ref{equ:flame1}), where the encoder pre-trained on abundant macro-expression data captures prior knowledge of facial expressions. (2) The flow extractor computes  the optical flow sequence $\{\mathcal{O}_t \in \mathbb{R}^{H \times W \times 2}\}_{t=1}^{T-1}$ between adjacent frames $I_t$ and $I_{t+1}$, and the motion encoder $E^{\text {motion}}$ captures the subtle temporal dynamic  $\Delta\psi_t$ of micro-expressions, given by 
\begin{equation}
     \Delta\psi_t = E^{\text {motion}}(\{\mathcal{O}_t\}_{t=1}^{T-1}). 
\end{equation}
We then bridge the two streams by designing a residual fusion mechanism $E^{\text {res}}$ that fuses the dynamics into the static reference, and obtain micro-expression features with global dynamics, given by 
\begin{equation}
    \psi_t = E^{\text {res}}(\psi_1, \Delta\psi_t, t). 
\end{equation}
The final output $\mathcal{M}_t^{\text {init}}$ combines shape, pose, and dynamically-enhanced expression parameters through the FLAME model in Eq. (\ref{equ:flame2}). 
\noindent\textbf{Motion Encoder:} 
% The Motion Encoder is designed to extract subtle temporal dynamics from optical flow sequences for micro-expression modeling. At its core is a 3D convolutional neural network (3D CNN) that performs convolution operations across both spatial and temporal dimensions, effectively capturing motion information between adjacent frames.
Given an input optical flow sequence $\{\mathcal{O}_t\}_{t=1}^{T-1}$, the motion encoder uses multiple layers of 3D convolution and downsampling to progressively extract spatio-temporal feature maps. The encoder ultimately outputs a sequence of residual expression parameters $\{\Delta \psi_t\}_{t=1}^{T-1}$.
% which includes dynamic variations for facial expressions ($\Delta \psi_t^{\text {exp}}$), eyelids ($\Delta \psi_t^{\text {eyelid}}$), and jaw ($\Delta \psi_t^{\text {jaw}}$). These residuals represent minute changes relative to the expression state of the initial frame (onset frame), which is crucial for capturing low-intensity micro-expressions.

% \subsubsection{}
\noindent\textbf{Residual Fusion:} 
We integrate the computed residual expression parameters $\{\Delta \psi_t\}_{t=1}^{T-1}$ into the facial expression  parameters $\psi_1$ of $I_1$ with  global dynamics, and perform residual fusion  to operate dynamic integration in a learned latent space. %We model expression evolution as a continuous transformation process, enabling more sophisticated and physiologically plausible facial animation.
The fusion process begins by projecting the reference expression parameters $\psi_1$ into a latent representation $z_0 = E^{\text {latent}}(\psi_1)$, where  $E^{\text {latent}}$ is an encoding network that maps the expression parameters to a lower-dimensional latent space. In this latent space, we use a neural ordinary differential equation (ODE)~\cite{RubanovaCD19} to model the continuous evolution of expressions, given by
% \begin{equation}
%     \frac{dz_t}{dt} = E^{\text {ode}} (z_t, \Delta\psi_t, t). 
% \end{equation}
% The solution provides the evolved latent state via
% \begin{equation}
%     z_t = z_0 + \int_0^t E^{\text {ode}}(z_\tau, \Delta\psi_\tau, \tau) d\tau
% \end{equation}
% where $E^{\text {ode}}$ is a neural network that learns the dynamics of expression changes. The final expression parameters are obtained by decoding the evolved latent state $\psi_t = D^{\text {latent}}(z_t)$, where $D^{\text {latent}}$ reconstructs the expression parameters from the latent representation. The complete fusion process can be  represented as
\begin{equation}
    \psi_t = D^{\text {latent}}\left(z_0 + \int_0^t E^{\text {ode}}(E^{\text {latent}}(\psi_1), \Delta\psi_\tau, \tau) d\tau\right), \label{equ:ode}
\end{equation}
where $E^{\text {ode}}$ is a neural network, and $D^{\text {latent}}$ reconstructs the expression parameters from the latent representation.
\begin{figure*}[t]
    \centering
    \includegraphics[width=0.85\linewidth]{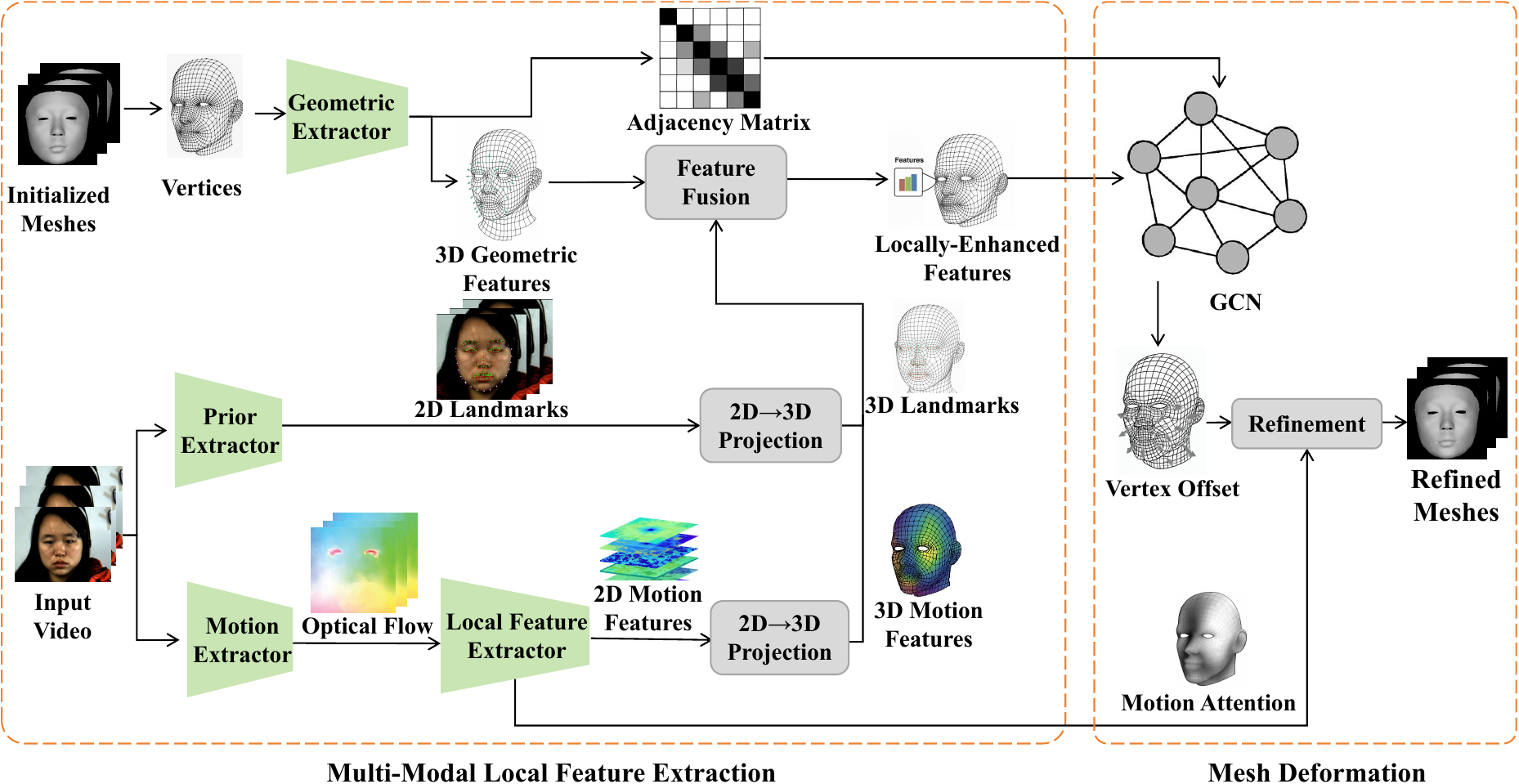}
    \caption{The \textit{dynamic-guided mesh deformation}  module refines the initialized meshes using locally-enriched features to reconstruct subtle dynamic details of  micro-expressions.}
    \label{fig:mesh_deformation}
\end{figure*}

\subsection{Dynamic-Guided Mesh Deformation}
\label{subsec:mesh_deformation}
%The dynamic-guided mesh deformation module aims to refine the initial 3D facial meshes by incorporating locally-enriched features to capture fine-grained details of micro-expressions. 
The dynamic-guided mesh deformation module operates on the initialized meshes $\{\mathcal{M}_t^{\text{init}}\}_{t=1}^T$ and deforms them into the refined meshes $\{\mathcal{M}_t^{\text{final}}\}_{t=1}^T$ through the multi-modal local feature extraction and graph-based mesh deformation, as shown in Figure~\ref{fig:mesh_deformation}. The refinement process is % can be formulated as
\begin{equation}
\mathcal{M}_t^{\text{final}} = \mathcal{G}(\mathcal{M}_t^{\text{init}}, E^{\text{local}}(I_t, \mathcal{O}_t,\mathcal{M}_t^{\text{init}})),
\end{equation}
where $\mathcal{G}$ represents the graph-based mesh deformation network, and $E^{\text{local}}$ denotes the multi-modal feature extraction. % from the input image $I_t$ and optical flow $\mathcal{O}_t$.

\subsubsection{Multi-Modal Local Feature Extraction}
% To capture comprehensive facial dynamics.
We extract features from multiple cues, yielding more stable and discriminative features of micro-expressions. The motivation is that features from multiple cues capture complementary information, where 3D geometric features preserve spatial consistency, facial priors from landmarks constrain plausible facial deformations, and 2D motion features capture subtle temporal variations.  

\noindent\textbf{3D Geometric Features:} 
We construct the adjacency matrix  and extract 3D geometric features $f^{\text{geo}}_t$ from the initialized mesh $\mathcal{M}_t^{\text{init}}$, given by
\begin{align}
A_t &= \mathcal{A}(\mathcal{M}_t^{\text{init}}),\\
f^{\text{geo}}_t &= E^{\text{geo}}(\mathcal{V}_t^{\text{init}},A_t).\label{equ:adj}
\end{align}
The adjacency matrix $A_t$ is defined from the graph structure of the initialized meshes $\mathcal{M}_t^{\text{init}}$, which  captures the topological relationships between vertices. The geometric encoder $E^{\text{geo}}$ processes the vertex coordinates through a multi-layer graph convolutional network to extract hierarchical spatial features with the   adjacency matrix $A_t$. 
% It operates directly on the mesh graph structure defined by the adjacency matrix $A$, which is constructed from mesh faces using a sparse representation that captures the topological relationships between vertices.

% The geometric feature extraction employs graph convolution layers that propagate information through the mesh connectivity:

% \begin{equation}
% F_{\text{geo}}^{(l+1)} = \sigma\left(\tilde{D}^{-\frac{1}{2}}\tilde{A}\tilde{D}^{-\frac{1}{2}}F_{\text{geo}}^{(l)}W^{(l)}\right)
% \end{equation}

% where $\tilde{A} = A + I$ adds self-connections, $\tilde{D}$ is the degree matrix, and $W^{(l)}$ are learnable weights at layer $l$. This geometric foundation ensures spatial coherence during the refinement process by preserving the intrinsic manifold structure of the facial surface while enabling local feature propagation across connected regions.

\noindent\textbf{Landmark Features:}
We use two types of facial landmarks to provide strong facial priors. (1) FAN landmarks $l_t^{\text{FAN}}$ with 68 facial keypoints detected using the face alignment network~\cite{BulatT17}, provide  comprehensive facial structure coverage. (2) MediaPipe landmarks $l_t^{\text{MP}} $~\cite{KartynnikAGG19} offer  enhanced details in critical regions like lips and eyes. The 3D landmark features $f_t^{\text{3D\_ldm}}$ are computed by
\begin{align}
% l_t^{\text{FAN}} &= E^{\text{FAN}}(I_t), \\
% l_t^{\text{MP}} &= E^{\text{MediaPipe}}(I_t), \\
% F_{\text{landmark}}^{2D} &= E_{\text{feat}}^{2D}(l_t^{\text{FAN}} \oplus l_t^{\text{MP}}), \\
f_t^{\text{3D\_ldm}}= \Pi^{-1}(l_t^{\text{FAN}} \oplus l_t^{\text{MP}}, \mathcal{K}),
\end{align}
where the projection $\Pi^{-1}$ uses camera parameters $\mathcal{K}$ to map 2D landmarks to  3D mesh space.  
% through orthographic projection:
% \begin{equation}
% \mathbf{v}_{\text{3D}} = \Phi_{\text{project}}(\mathbf{v}_{\text{2D}}, \mathbf{cam}) = \left[ \frac{2(\mathbf{v}_{\text{2D}}^x - t_x)}{s \cdot W} - 1, \ -\left(\frac{2(\mathbf{v}_{\text{2D}}^y - t_y)}{s \cdot H} - 1\right), \ 0 \right]
% \end{equation}
The projection ensures anatomical consistency by constraining deformations to physiologically plausible facial configurations while providing strong semantic guidance for local refinement.

\noindent\textbf{Motion-based Features:}
Motion-based features capture subtle temporal dynamics by processing dense optical flow through a specialized   convolutional neural network
\begin{align}
f_t^{\text{pixel\_motion}} = CNN^{\text{motion}}(\mathcal{O}_t), 
\end{align}
where $CNN^{\text{motion}}$ with spatial attention processes the optical flow to extract pixel-level features. 
To project these pixel-level features to 3D mesh vertices $\mathbf{v}$ effectively, we design an accelerated correspondence strategy that addresses the computational bottleneck of per-vertex projection.  Mapping all optical flow pixels to 3D space is computationally inefficient, and we observe that it is not necessary to calculate the 3D motion features of all optical-flow pixels when modeling micro-expressions. To this end, we develop an accelerated region-based pixel-vertex correspondence strategy that significantly reduces computational complexity while maintaining feature discriminability. 

% To optimize performance, we employ a multi-resolution acceleration strategy with two complementary approaches:

% \noindent\textbf{Grid-based Fast Mode:}
% For balanced accuracy and speed, we use a regular grid sampling approach:
% \begin{equation}
% \mathbf{G} = \{(x_i, y_j) | x_i = i \cdot \frac{W}{R_{\text{grid}}}, y_j = j \cdot \frac{H}{R_{\text{grid}}}, i,j \in [0, R_{\text{grid}}-1]\}
% \end{equation}
% where $R_{\text{grid}}$ controls the grid resolution. Vertex features are then computed via inverse distance weighting from grid samples:
% \begin{equation}
% F_{\text{motion}}^{3D}(v) = \frac{\sum_{\mathbf{g} \in \mathcal{N}(v)} w(\mathbf{g}, v) \cdot \text{bilinear\_sample}(F_{\text{motion}}^{\text{pixel}}, \mathbf{g})}{\sum_{\mathbf{g} \in \mathcal{N}(v)} w(\mathbf{g}, v)}
% \end{equation}
% with $w(\mathbf{g}, v) = \frac{1}{\|\Pi(v, \mathbf{cam}) - \mathbf{g}\| + \epsilon}$.
% \noindent\textbf{Region-based Approximation:}

% Specifically, the facial surface is partitioned into semantically meaningful regions based on anatomical structure: $\mathcal{R} =$ $ \{\text{eye\_left}, \text{eye\_right}, \text{nose}, \text{mouth}, \text{cheek\_left}, \text{cheek\_right}, $ $\text{forehead}, \text{chin}\}$. Each region $r \in \mathcal{R}$ is defined by a set of vertices $\mathcal{V}_r \subset \mathcal{V}_t^{\text{init}}$ that share similar motion characteristics during micro-expressions. For computational efficiency, we represent each region by a small patch $\mathcal{P}_r$ centered around its projected centroid in the 2D image space:

Specifically, the facial surface is partitioned into eight semantically meaningful regions $\mathcal{R}$ based on anatomical structure: left eye, right eye, nose, mouth, left cheek, right cheek, forehead, and chin. Each region $r \in \mathcal{R}$ contains a set of vertices $\mathcal{V}_r$ that exhibit similar motion patterns during facial expressions. %We define a function $\text{region}(v)$ maps each vertex $v$ to its corresponding semantic region $r$ based on the pre-defined facial partitioning 
To efficiently compute motion features, we determine the representative 2D position for each region by calculating its centroid $\mathbf{c}_r$, given by
% This is achieved by projecting all vertices $v \in \mathcal{V}_r$ from the 3D mesh to the 2D image space using the camera projection function $\Pi(v, \mathbf{cam})$, then computing their average position
\begin{equation}
\mathbf{c}_r = \frac{1}{|\mathcal{V}_r|} \sum_{\mathbf{v} \in \mathcal{V}_r} \Pi(\mathbf{v}, \mathcal{K}),
\end{equation}
where $\Pi(\mathbf{v}, \mathcal{K})$ denotes  projecting all vertices $\mathbf{v} \in \mathcal{V}_r$ from the 3D mesh to the 2D image space. 
Around each centroid $\mathbf{c}_r$, we extract a small $5 \times 5$ pixel patch $\mathcal{P}_r$ from the motion feature map. This patch $\mathcal{P}_r$ captures the local motion context surrounding the region's representative position. The region-level motion feature is computed by 
\begin{equation}
f_t^{\text{3D\_motion}} (\mathbf{v}) = \frac{1}{|\mathcal{P}_r|} \sum_{\mathbf{p} \in \mathcal{P}_r} f_t^{\text{pixel\_motion}}(\mathbf{p}), \label{equ:fast}
\end{equation}
where $r \in \mathcal{R}$ and $\mathbf{p}$ denotes the pixel. Eq. (\ref{equ:fast}) assigns the same region-level motion feature $f_t^{\text{3D\_motion}}$ to every vertex $\mathbf{v}$ belonging to the same facial region $r$.

% This formulation provides three key advantages: (1) \textbf{Computational Efficiency}: reducing the number of projection operations from $|\mathcal{V}|$ (typically 5023 vertices) to $|\mathcal{R}|$ (8 regions), achieving approximately 600$\times$ reduction in projection complexity; (2) \textbf{Noise Robustness}: spatial averaging within each region patch suppresses high-frequency noise while preserving meaningful motion patterns; and (3) \textbf{Semantic Coherence}: grouping vertices by anatomical regions ensures that vertices with similar functional roles receive consistent motion features.

% The region assignment function $\text{region}(v)$ is precomputed based on vertex coordinates in the canonical FLAME mesh topology, with regions defined by coordinate ranges in the facial parameter space. This static partitioning aligns with the functional units of facial action coding system (FACS), making it particularly suitable for capturing the localized dynamics of micro-expressions.

% Eq.(\ref{equ:fast}) dynamically selects the acceleration mode based on performance requirements, providing up to 5$\times$ speedup while maintaining over 90\% accuracy. This efficient approach allows the model to focus refinement on regions exhibiting significant micro-expression activity while maintaining real-time performance for practical applications.
\noindent\textbf{Feature Fusion:} 
The local features are fused to obtain locally-enhanced features that capture complementary information from different modalities. The fusion begins with feature concatenation followed by a fusion network
\begin{equation}
f_t^{\text{local}} = MLP^{\text{fusion}}(f_t^{\text{geo}} \oplus f_t^{\text{3D\_ldm}} \oplus f_t^{\text {3D\_motion}})
\end{equation}
The fusion network $MLP^{\text{fusion}}$ implements a multi-layer perceptron to effectively combine complementary information. 

\subsubsection{Mesh Deformation}
We use a graph convolutional network (GCN) to process the fused local features and predict vertex-wise displacements. The GCN operates directly on the mesh graph structure, propagating information through the facial topology defined by the adjacency matrix $A_t$. The deformation network takes  the concatenation of vertex coordinates and local features as input, 
\begin{equation}
\mathbf{X}^{(0)}_t = [\mathcal{V}_t^{\text{init}}; f_t^{\text{local}}].
\end{equation}
The graph convolution layers then iteratively update vertex features through neighborhood aggregation by
\begin{equation}
\mathbf{X}^{(l+1)}_t = \sigma\left(\tilde{D}^{-\frac{1}{2}}\tilde{A}_t\tilde{D}^{-\frac{1}{2}}\mathbf{X}^{(l)}_tW^{(l)}\right),
\end{equation}
where $\tilde{A}_t = A_t + I$ is the adjacency matrix with self-connections, $\tilde{D}$ is the degree matrix, $W^{(l)}$ denotes learnable weight matrices, and $\sigma$ is the ReLU activation function. After $L$ graph convolution layers, the network predicts vertex displacements through an output $MLP$, 
\begin{equation}
\Delta \mathcal{V}_t^{\text{base}} = MLP^{\text{output}}(\mathbf{X}^{(L)}_t). 
\end{equation}

% To maintain mesh quality during deformation, we incorporate several geometric regularization terms computed from the graph structure:

% \begin{align}
% \mathcal{L}_{\text{laplacian}} &= \|L\mathbf{V}\|_2^2 \quad &\text{(Surface Smoothness)} \\
% \mathcal{L}_{\text{edge}} &= \frac{1}{|E|}\sum_{e \in E} (\|e\| - \bar{l})^2 \quad &\text{(Edge Length Preservation)} \\
% \mathcal{L}_{\text{normal}} &= \frac{1}{|F|}\sum_{f \in F} (1 - \mathbf{n}_f \cdot \mathbf{n}_{f_0}) \quad &\text{(Normal Consistency)}
% \end{align}

% where $L$ is the Laplace-Beltrami operator, $E$ is the set of mesh edges, $F$ is the set of faces, and $\mathbf{n}_f$ are face normals.

\noindent\textbf{Motion-Attentive Vertex Refinement:}
To focus on regions exhibiting micro-expression activity, we introduce a motion attention mechanism that adaptively modulates vertex displacements based on optical flow intensity. This mechanism ensures that regions with significant motion receive more refinement while maintaining stability in relatively static areas.

% The refinement process begins by computing attention weights through an efficient region-based method. 
For each vertex $\mathbf{v}$, the attention weight $\mathbf{W}^{\text{attn}}(\mathbf{v})$ is determined by comparing the flow intensity in its corresponding region against an adaptive threshold. The regional flow intensity $I_t^{\text{flow}}(r)$ is computed as the average magnitude of optical flow within the $5 \times 5$ pixel patch $\mathcal{P}_r$ associated with region $r$, given by 
\begin{equation}
I_t^{\text{flow}}(r) = \frac{1}{|\mathcal{P}_r|} \sum_{\mathbf{p} \in \mathcal{P}_r} \|\mathcal{O}_t(\mathbf{p})\|,
\end{equation}
where $\|\mathcal{O}_t(\mathbf{p})\|$ is the flow magnitude at pixel $\mathbf{p}$ in frame $t$. Using this regional intensity measure, the attention weight for each vertex $\mathbf{v}$ located in the region $r$ is calculated as
\begin{equation}
\mathbf{W}^{\text{attn}}(\mathbf{v}) = \text{sigmoid}\left(\frac{I_t^{\text{flow}}(r) - \tau^{\text{adaptive}}}{\tau^{\text{adaptive}} + 10^{-4}}\right),
\end{equation}
where $\tau^{\text{adaptive}}$ is a dynamic threshold computed per frame using the statistical properties of flow intensity across the entire face, 
\begin{equation}
\tau^{\text{adaptive}} = \mu^{\text{flow}} + 0.5 \sigma^{\text{flow}}.
\end{equation}
$\mu^{\text{flow}}$ and $\sigma^{\text{flow}}$ denote the mean and standard deviation of flow intensity across all facial regions, respectively. This adaptive thresholding ensures that the attention mechanism responds appropriately to varying expression intensities across different video sequences.

The computed attention weights are then applied to modulate the base vertex displacements predicted by the graph convolutional network,
\begin{equation}
\Delta \mathcal{V}_t = \mathbf{W}^{\text{attn}} \odot (\lambda^{\text{attn}} \Delta \mathcal{V}_t^{\text{base}}),
\end{equation}
where $\odot$ denotes element-wise multiplication and $\lambda^{\text{attn}} = 0.02$ is a scaling factor that controls the overall magnitude of deformation. This scaling ensures subtle adjustments appropriate for capturing the low-intensity dynamics of micro-expressions while preventing excessive mesh distortion.

The final refined mesh is obtained by applying these motion-attentive vertex offsets to the initialized mesh
\begin{equation}
\mathcal{M}_t^{\text{final}} = (\mathcal{V}_t^{\text{init}} + \Delta \mathcal{V}_t, \mathcal{F}_t).
\end{equation}

\subsection{Optimization Objectives}
\label{subsec:optimization}
Our method uses the reconstruction fidelity loss $\mathcal{L}^{\text{rec}}$ and geometric regularization loss $\mathcal{L}^{\text{geo}}$ for training, given by
\begin{equation}
\mathcal{L}^{\text{total}} = \mathcal{L}^{\text{rec}}  + \lambda^{\text{geo}} \mathcal{L}^{\text{geo}},
\end{equation}
where $\lambda^{\text{geo}}$ denotes the trade-off parameter.

\noindent\textbf{Reconstruction Fidelity Loss:}
We use the reconstruction loss between the input frame $I_t$ and the rendered frame $\hat{I}_t$ in Eq. (\ref{equ:render}) to ensure photometric and perceptual fidelity, including the photometric loss $\mathcal{L}^{\text{photo}}$, VGG perceptual loss $\mathcal{L}^{\text{vgg}}$, landmark loss $\mathcal{L}^{\text{lmk}}$, expression regularization loss $\mathcal{L}^{\text{reg}}$, emotion loss $\mathcal{L}^{\text{emo}}$, expression consistency loss $\mathcal{L}^{\text{exp}}$, and identity consistency loss $\mathcal{L}^{\text{id}}$. For further details, we refer readers to \textit{supplementary materials}. % or the work of \cite{RetsinasFDARBM24}. 

\noindent\textbf{Geometry Regularization Loss:}
To maintain mesh quality during refinement, we use the geometry regularization loss $\lambda^{\text{geo}}$ operating directly on the mesh structure and its deformation,
\begin{equation}
\mathcal{L}^{\text{geo}} = \lambda^{\text{lap}} \mathcal{L}^{\text{lap}} + \lambda^{\text{normal}} \mathcal{L}^{\text{normal}} + \lambda^{\text{flow\_guide}} \mathcal{L}^{\text{flow\_guide}},
\end{equation}
where $\mathcal{L}^{\text{lap}}$, $\mathcal{L}^{\text{normal}}$ and $\mathcal{L}^{\text{flow\_guide}}$ represent the Laplacian smoothness loss, normal consistency loss and flow-guided local refinement loss, respectively. $ \lambda^{\text{lap}}$, $\lambda^{\text{normal}}$ and $\lambda^{\text{flow\_guide}}$ are their trade-off parameters. 

The Laplacian smoothness loss encourages smooth surface deformation by minimizing the Laplacian of the mesh displacement, given by
\begin{equation}
\mathcal{L}^{\text{lap}} =\mathbb{E} [\|L\Delta \mathcal{V}_t\|_2^2],
\end{equation}
where $L$ is the Laplace-Beltrami operator.
% constructed from the mesh adjacency matrix $A$ calculated in Eq. (\ref{equ:adj}).  $\mathcal{L}^{\text{lap}}$ penalizes high-frequency deformations, ensuring smooth surface transitions that are physiologically plausible for facial actions.

% The edge length preservation loss maintains uniform edge lengths throughout the deformation process:
% \begin{equation}
% \mathcal{L}^{\text{edge}} = \frac{1}{|E|}\sum_{e \in E} (\|e\| - \bar{l})^2,
% \end{equation}
% where $E$ is the set of all edges in the mesh, $\|e\|$ is the length of edge $e$ in the refined mesh $\mathcal{M}_t^{\text{final}}$, and $\bar{l}$ is the mean edge length computed from the initial mesh $\mathcal{M}_t^{\text{init}}$. This regularization prevents excessive stretching or compression of mesh elements during refinement.
The normal consistency loss preserves local surface orientation by maintaining face normal coherence
\begin{equation}
\mathcal{L}^{\text{normal}} = \frac{1}{|\mathcal{F}|}\sum_{f \in \mathcal{F}} (1 - \mathbf{n}_f  \mathbf{n}_{f_0}^{\text T}),
\end{equation}
where $\mathcal{F}$ is the set of all triangular faces in the mesh, $\mathbf{n}_f$ is the unit normal of the face $f$ in the refined mesh, and $\mathbf{n}_{f_0}$ is the corresponding face normal in the initialized mesh.
% $\mathcal{L}^{\text{normal}}$ ensures that local surface characteristics are preserved during deformation.

The flow-guided local refinement loss encourages appropriate vertex displacement based on optical flow intensity
\begin{equation}
% \mathcal{L}^{\text{flow\_guide}} = \mathbb{E}[\|\Delta \mathcal{V}_t\| \cdot \mathbf{W}^{\text{attn}}] + \lambda^{\text{low}}\mathbb{E}[\|\Delta \mathcal{V}_t\| \cdot \mathbf{M}^{\text{low}}],\label{equ:flowguide}
\mathcal{L}^{\text{flow\_guide}} = \mathbb{E}[\|\Delta \mathcal{V}_t\|\mathbf{M}^{\text{low}}] -\lambda^{\text{flow}}\mathbb{E}[\|\Delta \mathcal{V}_t\| \mathbf{W}^{\text{attn}}],\label{equ:flowguide}
\end{equation}
where $\mathbf{M}^{\text{low}}$ is a binary mask that identifies low-flow regions where $\mathbf{W}^{\text{attn}}(\mathbf{v}) < 0.5$, and $\lambda^{\text{flow}}=0.1$ is the trade-off parameter.
% The first term of Eq. (\ref{equ:flowguide}) encourages larger displacements in high-flow regions, while the second term enforces sparsity by penalizing large displacements in low-flow areas.

% These geometric regularization terms work together to ensure that our mesh refinement produces physically plausible deformations while preserving the intrinsic quality of the facial mesh structure.
\begin{figure*}[t]
    \centering
    \includegraphics[width=0.75\linewidth]{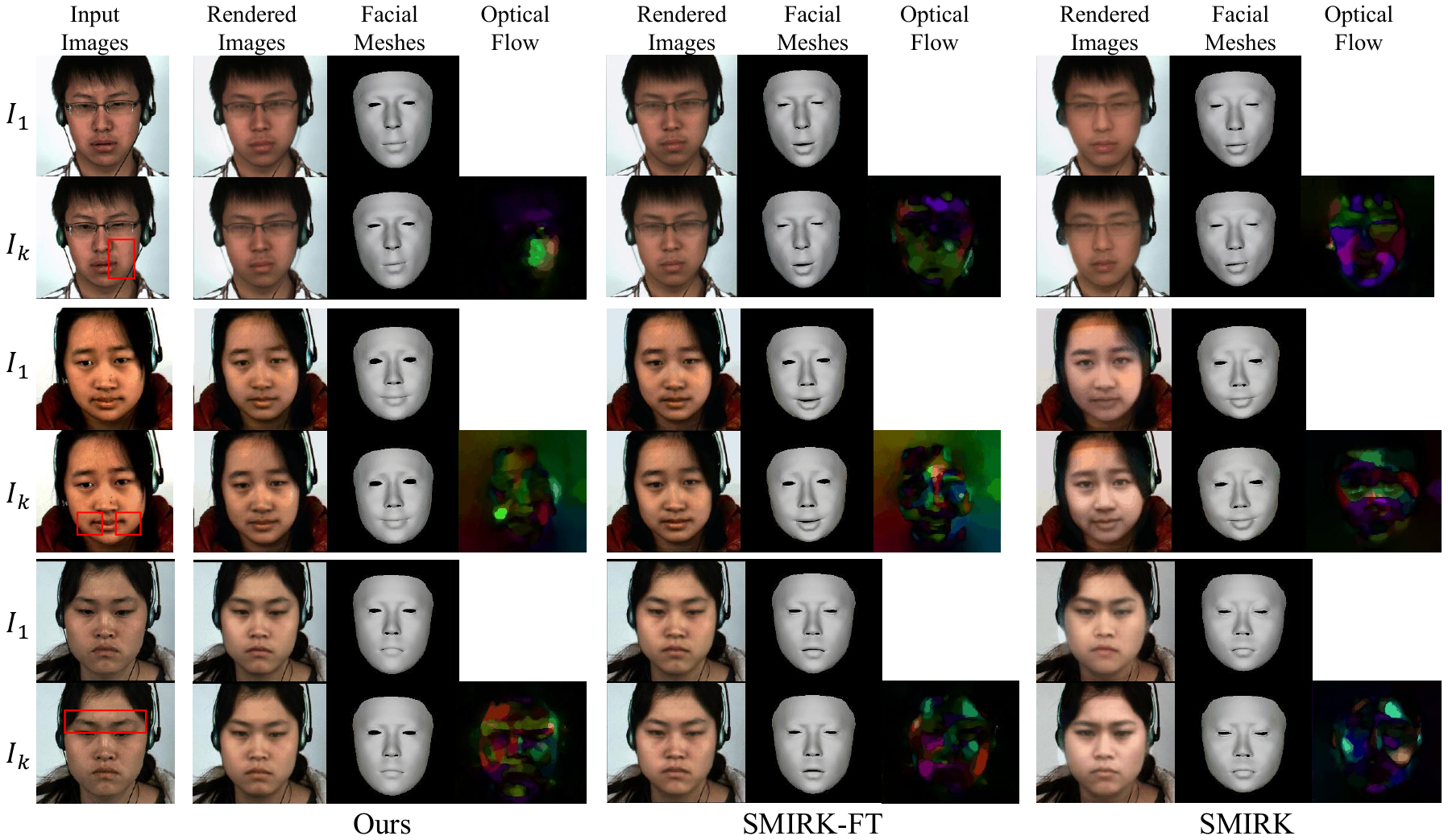}
    \caption{Visualizations of our method, SMIRK with micro-expression data fine-tuning (SMIRK-FT), and SMIRK without fine-tuning. The red boxes highlight the facial regions where noticeable micro-expression changes occur at $I_k$.}
    \label{fig:vis}
\end{figure*}

\section{Experiments}
\label{sec:experiments}

\subsection{Datasets}
\label{subsec:datasets}
% We evaluate our method on three widely used micro-expression datasets: CASME~\cite{yan2013casme}, CASME II~\cite{YanLWZLCF14}, and SAMM~\cite{DavisonLCTY18}.
We repurpose three high-frame-rate micro-expression recognition datasets, namely CASME~\cite{yan2013casme}, CASME II~\cite{YanLWZLCF14}, and SAMM~\cite{DavisonLCTY18}, for evaluating our method. These datasets have demonstrated the ability to capture subtle and rapid facial dynamics that are essential for accurate micro-expression reconstruction. CASME, the pioneering dataset in the CASME series, contains 195 micro-expression videos captured from 26 subjects at 100 fps with natural emotion induction. CASME II contains 255 micro-expression videos captured at 200 fps. SAMM consists of 159 videos recorded at 200 fps under controlled lighting conditions. For each dataset, we randomly select 70\% of the videos for training and 30\% for testing. %SMIC includes 164 videos captured at 100 fps. %For all datasets, we process each micro-expression clip by uniformly sampling 8 frames from the onset to offset frames to capture the complete temporal evolution while maintaining computational efficiency. This sampling strategy ensures capturing the subtle dynamics throughout the micro-expression duration while reducing redundancy in high-frame-rate sequences.

% \textbf{CASME II} contains 255 micro-expression videos from 26 subjects, recorded at 200 fps with a resolution of $280 \times 340$ pixels. The dataset includes five main expression categories: happiness, disgust, repression, surprise, and others for five-class evaluation, and positive, negative, and surprise for three-class evaluation.

% \textbf{SAMM} consists of 159 micro-expression videos from 29 subjects, captured by a grayscale camera at 200 fps under controlled lighting conditions. We use happiness, anger, contempt, surprise, and others for five-class evaluation, and positive, negative, and surprise for three-class evaluation.

% \textbf{SMIC} includes 164 micro-expression videos from 16 subjects, recorded at 100 fps and labeled with three emotion classes: positive, negative, and surprise.

% Since 3D ground truth annotations are not available in these datasets, we employ indirect evaluation metrics to assess reconstruction quality, including geometric accuracy, perceptual fidelity, and downstream task performance.

\subsection{Implementation Details}

% Our Dynamic-Encoded Module with a 3D CNN motion encoder (5 layers, channels [64,128,256,512,512]) and neural ODE fusion; (2) \textit{Dynamic-Guided Mesh Deformation} with a 4-layer GCN (dims [128,256,512,256]) and multi-modal fusion MLP.

% We use a MobileNetV3 backbone to construct the static encoder. The motion encoder utilizes 5 layers of 3D convolution. Both the neural ODE in Eq. (\ref{equ:ode})  and the multi-modal fusion MLP use a 3-layer MLP.  
%The graph convolutional network employs 2 graph convolution layers with feature dimensions [128, 256, 512, 256]. The multi-modal fusion MLP has architecture [384, 512, 256, 128] for combining geometric, landmark, and motion features. 
We uniformly sample 8 frames from the onset to offset of each micro-expression clip to capture complete temporal dynamics while reducing redundancy. Frames are aligned and resized to $224\times224$. Following SMIRK~\cite{RetsinasFDARBM24}'s analysis-by-synthesis framework, we first pre-train the static encoder, render and augmented expression cycle path on four macro-expression datasets of FFHQ~\cite{karras2019style}, CelebA~\cite{liu2015deep}, LRS3~\cite{afouras2018lrs3}, and MEAD~\cite{wang2020mead}. For micro-expression fine-tuning, we only optimize the dynamic-encoded module and dynamic-guided mesh deformation module using the Adam optimizer~\cite{kingma2014adam} with a learning rate $1e-4$ and batch size 4. Optical flow between frames is computed using TV-L1~\cite{zach2007duality}. Additional implementation details are provided in the \textit{supplementary material}.

\subsection{Evaluation Metrics}
\label{subsec:metrics}

Inspired by the work of SMIRK~\cite{RetsinasFDARBM24}, we use the following metrics to evaluate our method. %Different from SMIRK which compare L1 and VGG losses for evaluation, we focus on evaluation metrics that directly measure the quality of micro-expression reconstruction. 
% \noindent\textbf{MER Accuracy: } 
% \textit{MER Accuracy: } 
 We evaluate the perceptual quality of rendered micro-expressions in Eq. (\ref{equ:render}) by measuring the performance of micro-expression recognition using a pre-trained MOL classifier~\cite{ShaoCLZLXM25}, and report accuracy (\textit{Acc}) and weighted F1-score (\textit{WF1}) for  evaluation. %and Unweighted F1-score (UF1) and Unweighted Average Recall (UAR) for composite dataset evaluation. The reconstructed 3D mesh sequences are rendered into image sequences and fed into the classifier, with higher scores indicating better preservation of subtle expression dynamics crucial for recognition tasks.
% \noindent\textbf{Landmark Reprojection Error (LRE): } 
% We compute the mean Euclidean distance between detected 2D facial landmarks and the projected 3D mesh landmarks. 
% \[
% \text{LRE} = \frac{1}{T} \sum_{t=1}^{T} \frac{1}{K} \sum_{k=1}^{K} \| \mathbf{l}_{t,k} - \Pi(\mathbf{v}_{t,k}) \|_2, 
% \]
% where $\mathbf{l}_{t,k}$ are ground truth 2D landmarks, $\mathbf{v}_{t,k}$ are corresponding 3D mesh vertices, and $\Pi$ is the projection function. Lower LRE values indicate better geometric accuracy.
% \noindent\textbf{L1 Loss and VGG Loss: } 
% \textit{L1 Loss and VGG Loss: } 
We also evaluate the faithfulness of 3D facial reconstruction by comparing the \textit{L1 Loss} and \textit{VGG Loss} between the rendered images and input images. 
% \noindent\textbf{FID Score: } 
% \textit{FID Score: } 
We use Fréchet Inception Distance (\textit{FID}) to measure the perceptual quality between the input images and rendered images. %Lower FID values indicate better preservation of visual details and expression characteristics in the reconstructed sequences.

% \noindent\textbf{User Study (Optional): } 
% Following~\cite{RetsinasFDARBM24}, we conduct a perceptual user study where participants compare reconstructed results from different methods. Users are presented with original videos alongside reconstructions and asked to rate the faithfulness of expression reproduction on a 5-point Likert scale.

\subsection{Quantitative Results}
\label{subsec:quantitative}

% \subsubsection{Comparison with State-of-the-Art}

We compare our method with state-of-the-art 3D facial expression reconstruction methods, including SMIRK~\cite{RetsinasFDARBM24}, EMOCA~\cite{DanecekBB22}, and EMICA~\cite{EMOTE}. To ensure a fair comparison under micro-expression scenarios, we extract the reconstructed meshes from EMOCA and EMICA, and introduce the neural render of the fine-tuned SMIRK (denoted as SMIRK-FT) using micro-expression datasets, to jointly generate the corresponding rendered images in Eq. (\ref{equ:render}).

% \begin{table}[h]
% \centering
% \small
% \setlength{\tabcolsep}{2pt}
% \caption{Quantitative comparison on micro-expression datasets. Best results are in \textbf{bold}.}
% \label{tab:quantitative_comparison}
% \begin{tabular}{@{}lcccccccccc@{}}
% \hline
% & \multicolumn{3}{c}{\textbf{CASME II}} & \multicolumn{3}{c}{\textbf{SAMM}} & \multicolumn{3}{c}{\textbf{SMIC}} & \textbf{Avg.} \\
% \cmidrule(lr){2-4} \cmidrule(lr){5-7} \cmidrule(lr){8-10} \cmidrule(lr){11-11}
% \textbf{Method} & \textbf{Acc} & \textbf{WF1} & \textbf{LRE} & \textbf{Acc} & \textbf{WF1} & \textbf{LRE} & \textbf{Acc} & \textbf{WF1} & \textbf{LRE} & \textbf{FID} \\
% \hline
% SMIRK & 68.3 & 66.5 & 3.45 & 65.7 & 63.8 & 3.62 & 70.1 & 68.9 & 3.28 & 45.5 \\
% EMOCA & 72.1 & 70.8 & 3.21 & 68.9 & 67.3 & 3.38 & 73.5 & 72.1 & 3.05 & 42.4 \\
% DECA & 65.8 & 63.9 & 3.89 & 63.2 & 61.5 & 4.05 & 67.9 & 65.7 & 3.72 & 48.9 \\
% \textbf{Ours} & \textbf{83.2} & \textbf{81.7} & \textbf{2.15} & \textbf{80.1} & \textbf{78.4} & \textbf{2.31} & \textbf{79.8} & \textbf{77.9} & \textbf{2.42} & \textbf{33.9} \\
% \hline
% \end{tabular}
% \end{table}

\begin{table}[t]
\centering
% \small
\setlength{\tabcolsep}{2.5pt}
\caption{Quantitative comparison on micro-expression datasets. Best results are in \textbf{bold}.}
\label{tab:quantitative_comparison}
\begin{tabular}{lcccc}
\hline
\textbf{Method} & \textbf{CASME II} & \textbf{CASME} & \textbf{SAMM} & \textbf{Avg.} \\
\hline
\textbf{ACC (\% $\uparrow$)} \\
EMOCA & 40.00 & 38.93 & 31.37 & 36.77 \\
EMICA & 42.50 & 28.81 & 29.41 & 33.57 \\
SMIRK & 35.00 & 44.07 & 45.10 & 41.39 \\
SMIRK-FT & 46.25 & 42.37 & 50.98 & 46.53 \\
\textbf{Ours} & \textbf{53.75} & \textbf{44.70} & \textbf{56.86} & \textbf{51.77} \\
\hline
\textbf{WF1 (\%  $\uparrow$)} \\
EMOCA & 38.05 & 28.32 & 19.74 & 28.70 \\
EMICA & 38.76 & 21.50 & 15.38 & 25.21 \\
SMIRK & 32.84 & 28.65 & 41.05 & 34.18 \\
SMIRK-FT & 46.35 & 25.22 & 49.51 & 40.36 \\
\textbf{Ours} & \textbf{53.32} & \textbf{28.65} & \textbf{54.58} & \textbf{45.52} \\
\hline
\textbf{L1 Loss $\downarrow$} \\
EMOCA & 0.085 & 0.196 & 0.346 & 0.209 \\
EMICA & 0.083 & 0.197 & 0.361 & 0.214 \\
SMIRK & 0.085 & 0.152 & 0.153 & 0.130 \\
SMIRK-FT & 0.050 & 0.051 & 0.069 & 0.057 \\
\textbf{Ours} & \textbf{0.041} & \textbf{0.044} & \textbf{0.060} & \textbf{0.048} \\
\hline
\textbf{VGG Loss $\downarrow$} \\
EMOCA & 1.578 & 1.302 & 1.589 & 1.490 \\
EMICA & 1.501 & 1.308 & 1.546 & 1.452 \\
SMIRK & 1.032 & 1.275 & 1.260 & 1.189 \\
SMIRK-FT & 0.745 & 0.567 & 0.796 & 0.703 \\
\textbf{Ours} & \textbf{0.700} & \textbf{0.539} & \textbf{0.742} & \textbf{0.660} \\
\hline
\textbf{FID $\downarrow$} \\
EMOCA & 112.37 & 412.20 & 448.69 & 324.42 \\
EMICA & 100.04 & 438.72 & 469.11 & 335.96 \\
SMIRK & 52.26 & 349.97 & 122.13 & 174.79\\
SMIRK-FT & 33.80 & 120.28 & 44.19 & 66.09 \\
\textbf{Ours} & \textbf{30.41} & \textbf{100.16} & \textbf{39.76} & \textbf{56.78} \\
\hline
\end{tabular}
\end{table}

% As shown in Table~\ref{tab:quantitative_comparison}, our method achieves superior performance across all metrics and datasets. The significant improvement in the MER accuracy (51.77\% average) and weighted F1 score (45.52\%) demonstrates that our reconstructions better preserve the subtle dynamics crucial for micro-expression recognition. For fair comparison, no model uses any information from the micro expression classifier \cite{ShaoCLZLXM25} during training. The result also indicates that we have extracted more discriminative micro-expression features.  Our method outperforms the second-best approach by 5.24\% in accuracy and 5.16\% in weighted F1-score, highlighting its effectiveness in capturing micro-expression nuances. The reconstruction accuracy, measured by L1 Loss and VGG Loss, shows our method achieves the lowest reconstruction errors (0.048 and 0.660 average), indicating successful refinement of local facial details while maintaining global consistency. In perceptual quality, our method achieves the best FID scores (56.78 average), suggesting more realistic reconstructions of micro-expressions. 

As shown in Table~\ref{tab:quantitative_comparison}, our method achieves superior performance across all metrics and datasets, demonstrating consistent improvements over existing facial reconstruction methods. In micro-expression recognition, our method achieves 51.77\% average accuracy (Acc) with a 5.24\% improvement over SMIRK-FT (46.53\%), particularly obtaining strong gains in CASME II (+7.50\%) and SAMM (+5.88\%). The weighted F1 score (WF1) similarly improves by 5.16\% on average. For reconstruction quality, our method reduces L1 loss by 0.009 (0.048 vs 0.057) and VGG loss by 0.043 (0.660 vs 0.703) on average compared with SMIRK-FT, indicating better detail preservation. In perceptual realism, our method achieves a 9.31 FID improvement (56.78 vs 66.09), with the most significant gain in CASME, compared with the top-performance method SMIRK-FT.

% These results validate our architecture's effectiveness in capturing subtle micro-expression dynamics while maintaining robust performance across diverse datasets.

In summary, our method demonstrates clear advantages in capturing subtle micro-expression dynamics while maintaining robust performance across diverse datasets.

\subsection{Qualitative Results}
\label{subsec:qualitative}
We visualize representative failure cases in Figure~\ref{fig:failure}, which shows the onset and peak frames (input images), rendered facial images and meshes, and the optical flow between input images. In case (a), global motion due to head jitter produces large and noisy optical flow across the entire face, overwhelming the subtle mouth movement and preventing its accurate reconstruction. In case (b), although the optical flow correctly captures eye motion, our FLAME-based mesh lacks detailed 3D modeling of the eye region (e.g., eyelid deformation or eyeball rotation), limiting expressiveness in this area. Cases (c) and (d) illustrate scenarios where the facial motion is extremely subtle. The resulting optical flow magnitude falls below the sensitivity threshold of standard flow estimators, leading to near-zero motion signals and consequently no detectable mesh deformation. % These examples highlight two key challenges: robustness to motion noise and sensitivity to ultra-low-amplitude dynamics—both of which motivate our future direction on learning micro-expression representations from imperfect motion cues.

% \begin{figure}[h]
% \centering
% \includegraphics[width=0.95\textwidth]{figures/qualitative_comparison.pdf}
% \caption{Visual comparison of 3D micro-expression reconstruction. From left to right: Input frame, SMIRK, EMOCA, DECA, Ours. Our method better captures subtle lip corner raising and eyebrow movements.}
% \label{fig:qualitative_comparison}
% \end{figure}

% Our method demonstrates superior temporal consistency, as shown in Figure~\ref{fig:temporal_consistency}. The expression parameters evolve smoothly while preserving the subtle nature of micro-expressions.

% \begin{figure}[h]
% \centering
% \includegraphics[width=0.8\textwidth]{figures/temporal_analysis.pdf}
% \caption{Temporal analysis of expression parameters. Our method shows smoother evolution while maintaining subtle dynamics.}
% \label{fig:temporal_consistency}
% \end{figure}

\begin{table}[t]
\centering
\small
\setlength{\tabcolsep}{2pt}
\caption{Comprehensive ablation studies on main components, multi-modal features, and loss functions.}
\label{tab:comprehensive_ablation}
\begin{tabular}{lccccc}
\hline
\textbf{Ablation Type} & \textbf{Acc (\%)} & \textbf{WF1 (\%)} & \textbf{L1 Loss} & \textbf{VGG Loss} & \textbf{FID} \\
\hline
\multicolumn{6}{l}{\textit{Main Components}} \\
\hline
w/o DEM & 46.25 & 46.07 & 0.042 & \textbf{0.695} & 30.78 \\
w/o DGMD & 47.25 & 47.28 & 0.043 & 0.724 & 31.70 \\
w/o Pre-Training & 50.00 & 49.85 & 0.076 & 0.867 & 45.95 \\
\hline
\multicolumn{6}{l}{\textit{Multi-modal Features}} \\
\hline
w/o Geometry & 46.25 & 45.52 & 0.042 & 0.704 & 31.85 \\
w/o Landmarks & 50.00 & 44.39 & 0.042 & 0.702 & 31.48\\
w/o Motion & 45.00 & 44.84 & 0.043 & 0.699 & 31.42 \\
\hline
\multicolumn{6}{l}{\textit{Loss Functions}} \\
\hline
w/o $\mathcal{L}^{\text{geo}}$ & 46.25 & 46.20 & 0.042 & 0.701 & 30.78 \\
w/o $\mathcal{L}^{\text{lap}}$ & 51.00 & 50.52 & 0.043 & 0.705 & 30.67 \\
w/o $\mathcal{L}^{\text{normal}}$ & 48.75 & 48.36 & 0.043 & 0.704 & 30.66\\
w/o $\mathcal{L}^{\text{flow\_guide}}$ & 51.25 & 50.65 & 0.041 & 0.700 & 30.70 \\
\hline
\textbf{Full Model} & \textbf{53.75} & \textbf{53.32} & \textbf{0.041} & 0.700 & \textbf{30.41} \\
\hline
\end{tabular}
\end{table}

\subsection{Ablation Studies}
\label{subsec:ablation}

We conduct comprehensive ablation studies to validate our designs. All experiments are performed on CASME II using the five-class evaluation protocol, and the results are shown in Table \ref{tab:comprehensive_ablation}.

\noindent\textbf{Component Analysis:} The ablation reveals distinct roles of each component. Removing the dynamic-encoded module  (``w/o DEM'') causes the largest accuracy drop (46.25\% vs 53.75\%) due to its role in temporal dynamics, and removing dynamic-guided mesh deformation (``w/o DGMD'') reduces accuracy by 6.50\%. The significant reconstruction quality degradation occurs without pre-training (``w/o Pre-Training'') from the evaluation of L1 Loss, VGG Loss and FID. This demonstrates that pre-training establishes robust facial modeling foundations, while our specialized modules enable precise micro-expression dynamics extraction.

% \begin{table}[t]
% \centering
% \small
% \setlength{\tabcolsep}{3pt}
% \caption{Ablation study on multi-modal features.}
% \label{tab:feature_ablation}
% \begin{tabular}{lccccc}
% \hline
% \textbf{Features} & \textbf{Acc (\%)} & \textbf{WF1 (\%)} & \textbf{L1 Loss}& \textbf{VGG Loss} & \textbf{FID} \\
% \hline
% w/o Geometry  & - & - & - & -& - \\
% w/o Landmarks & - & - & - & -& - \\
% w/o Motion & - & -& - & -& - \\
% \hline
% \textbf{All Features} & \textbf{53.75} & \textbf{53.32} & \textbf{0.041} & 0.700 & \textbf{30.41}\\
% \hline
% \end{tabular}
% \end{table}
% \noindent\textbf{Feature Analysis: }
% Table~\ref{tab:comprehensive_ablation} demonstrates the progressive improvement when combining multiple feature modalities. (1) Geometric features provide the spatial foundation but lack temporal dynamics information; (2) landmark features add strong anatomical priors, improving geometric accuracy and recognition performance; (3) optical flow contribute most significantly to capturing subtle temporal dynamics, with the largest performance gain among individual feature additions.
\noindent\textbf{Feature Analysis:} The significant performance drop when removing motion features (``w/o Motion'', 8.75\% accuracy decrease) underscores the critical role of dynamics. The removal of geometric and landmark features also reduces performance, confirming that each modality provides unique spatial and facial priors essential for robust reconstruction.

% \noindent\textbf{Feature Analysis:} Motion features contribute most significantly, with their removal causing an 8.75\% accuracy drop (45.00\% vs 53.75\%). Geometric and landmark features provide complementary spatial and anatomical priors for robust reconstruction.

% \subsubsection{}

% \begin{table}[t]
% \centering
% \small
% \setlength{\tabcolsep}{3pt}
% \caption{Ablation study on multiple losses.}
% \label{tab:loss_ablation}
% \begin{tabular}{lccccc}
% \hline
% \textbf{Losses} & \textbf{Acc (\%)} & \textbf{WF1 (\%)} & \textbf{L1 Loss}& \textbf{VGG Loss} & \textbf{FID} \\
% \hline
% w/o $\mathcal{L}^{\text{geo}}$ & - & - & - & -& - \\
% w/o $\mathcal{L}^{\text{lap}}$  & - & - & - & -& - \\
% w/o $\mathcal{L}^{\text{normal}}$ & - & - & - & -& - \\
% w/o $\mathcal{L}^{\text{flow\_guide}}$  & - & - & - & -& - \\
% \hline
% \textbf{All Losses} & \textbf{53.75} & \textbf{53.32} & \textbf{0.041} & 0.700 & \textbf{30.41}\\
% \hline
% \end{tabular}
% \end{table}
% \noindent\textbf{Loss Function Analysis: }
% Ablation experiments in Table~\ref{tab:comprehensive_ablation} verify each loss term's contribution by sequential removal. Each exclusion degrades relevant metrics, confirming the synergy of complementary losses for micro-expression reconstruction.
% \noindent\textbf{Loss Function Analysis:} Geometric regularization losses collectively contribute to performance, with $\mathcal{L}^{\text{geo}}$ removal causing a 7.50\% accuracy drop and $\mathcal{L}^{\text{normal}}$ removal causing a 5.00\% drop, confirming their importance for stable mesh deformation.
\noindent\textbf{Loss Function Analysis:} Geometric regularization losses collectively enhance performance, with $\mathcal{L}^{\text{geo}}$ removal causing 7.50\% accuracy drop.  $\mathcal{L}^{\text{normal}}$ preserves local details, while $\mathcal{L}^{\text{lap}}$ and $\mathcal{L}^{\text{flow\_guide}}$ ensure smooth deformation and motion-aware refinement, respectively.  Therefore, removing any of them reduces recognition accuracy, confirming their complementary roles. % in micro-expression reconstruction.

% \subsubsection{Key Insights}

% The ablation studies reveal several critical factors for micro-expression reconstruction:

% 1. \textbf{Global-Local Integration:} The combination of global dynamic features and locally-enriched features is essential for capturing both holistic motion patterns and subtle local deformations.

% 2. \textbf{Multi-modal Cues:} Optical flow provides crucial temporal information, while facial landmarks offer anatomical constraints for plausible deformations.

% 3. \textbf{Data Efficiency:} Leveraging macro-expression data through progressive pre-training effectively mitigates micro-expression data scarcity.

% 4. \textbf{Architecture Specialization:} Each component in our framework contributes uniquely to handling the low-intensity nature of micro-expressions.
\subsection{Limitation and Discussion}
While our method demonstrates good performance in reconstructing micro-expressions, it still presents two main limitations. First, although our region-based acceleration significantly improves efficiency, the per-vertex optimization remains computationally demanding and fails to achieve real-time performance. Since micro-expressions commonly involve only sparse facial regions, exploring sparse-region representations offers a promising direction for further work. Second, optical flow images are often affected by noise, and assigning excessive weight to $\mathcal{L}^{\text{flow\_guide}}$ may introduce mesh distortions. Developing more robust strategies to extract micro-expression cues from noisy optical flow constitutes an important avenue for future research. Please refer to the \textit{supplementary materials} for more results and analysis.

\section{Conclusion}
\label{sec:clu}
We have presented a fine-grained method for 3D facial micro-expression reconstruction from monocular videos. The proposed method can address the unique challenges posed by the subtle, transient, and low-intensity nature of micro-expressions. We have devised  a dynamic-encoded module that can leverage macro-expression data to alleviate the scarcity of micro-expression data, along with a dynamic-guided mesh deformation module that can fuse multi-modal local features for detail refinement. Extensive experiments on three datasets of CASME, CASME II, and SAMM can demonstrate the effectiveness of our method.

\bibliographystyle{IEEEtran}
\bibliography{main}

\newpage

 

% \end{IEEEbiography}



\vfill

\end{document}